\documentclass[conference]{IEEEtran}
\IEEEoverridecommandlockouts

\usepackage{cite}
\usepackage{amsmath,amssymb,amsfonts}
\usepackage{algorithmic}
\usepackage{graphicx}
\usepackage{textcomp}
\usepackage{xcolor}
\usepackage{booktabs} 
\usepackage{caption}  
\usepackage{adjustbox}
\def\BibTeX{{\rm B\kern-.05em{\sc i\kern-.025em b}\kern-.08em
    T\kern-.1667em\lower.7ex\hbox{E}\kern-.125emX}}
\begin{document}

\title{Disruptive Attacks on Face Swapping via Low-Frequency Perceptual Perturbations}
\author{Minglei Shu\textsuperscript{*} \thanks{\textsuperscript{*} Corresponding author: Minglei Shu (shuml@sdas.org)}}

\author{
\begin{tabular}{cc}
Mengxiao Huang & Minglei Shu\textsuperscript{*} \\
Department of Mathematics and Artificial Intelligence & Shandong Artificial Intelligence Institute \\
Qilu University of Technology & Qilu University of Technology \\
Jinan, China & Jinan, China \\
huangmx001104@163.com & shuml@sdas.org
\\[2ex]
Shuwang Zhou & Zhaoyang Liu \\
Shandong Artificial Intelligence Institute & Shandong Artificial Intelligence Institute \\
Qilu University of Technology & Qilu University of Technology \\
Jinan, China & Jinan, China \\
zhoushw@qlu.edu.cn & liuzhyang@qlu.edu.cn
\end{tabular}
\thanks{\textsuperscript{*}Corresponding author: Minglei Shu (shuml@sdas.org)}
}

\maketitle
\begin{abstract}
Deepfake technology, driven by Generative Adversarial Networks (GANs), poses significant risks to privacy and societal security. Existing detection methods are predominantly passive, focusing on post-event analysis without preventing attacks. To address this, we propose an active defense method based on low-frequency perceptual perturbations to disrupt face-swapping manipulation, reducing the performance and naturalness of generated content. Unlike prior approaches that used low-frequency perturbations to impact classification accuracy, our method directly targets the generative process of deepfake techniques.We combine frequency and spatial domain features to strengthen defenses. By introducing artifacts through low-frequency perturbations while preserving high-frequency details, we ensure the output remains visually plausible. Additionally, we design a complete architecture featuring an encoder, a perturbation generator, and a decoder, leveraging discrete wavelet transform (DWT) to extract low-frequency components and generate perturbations that disrupt facial manipulation models. Experiments on CelebA-HQ and LFW demonstrate significant reductions in face-swapping effectiveness, improved defense success rates, and preservation of visual quality.
\end{abstract}


\section{Introduction}
Deep forgery, distinct from traditional image tampering or earlier media manipulation, relies on AI models trained through deep learning on extensive datasets. Using techniques like Generative Adversarial Networks (GANs), attackers can create highly realistic synthetic content, making it difficult to distinguish real from fake media. While deep forgery has benign uses in film and television, it is also prone to abuse. Malicious actors can manipulate facial features, such as hair color, age, or appearance \cite{b1,b2,b3,b4,b5}, perform face-swapping \cite{b6,b7,b8,b9,b10}, or even achieve full facial reconstruction \cite{b11,b12}. These capabilities pose risks to privacy, spread misinformation, and damage reputations, particularly for public figures \cite{b13}. Consequently, deep forgery has drawn significant attention from researchers and policymakers due to its far-reaching societal implications.

Researchers have proposed various solutions to mitigate the harm caused by malicious Deepfake applications. Most current detection methods rely on passive analysis \cite{b21,b22,b23,b24,b25, b26}, using visual computing and machine learning to identify manipulation traces in images. However, these approaches often have low accuracy and poor generalization when confronting new Deepfake techniques.

Following this, many researchers have started with active defense by adding invisible information to the image before Deepfake \cite{b31,b34,b35,b42}, i.e., actively protecting the image before the generation of false content to stop the generation or dissemination of false information. Specifically, they block Deepfake generation by adding perturbations to the image to generate visually apparent artifacts. Most existing work on proactive defense against perturbations targets generative models that modify attributes \cite{b14,b15,b16,b17}. At the same time, TAFIM \cite{b18} embeds specific perturbations in the original image and combines them with attention-based specific perturbation fusion to protect personal privacy and information security. DF-RA \cite{b19} protects face images from Deepfake in various online social network (OSN) compression environments. However, their work is not satisfactory against face-swapping models.

To address these issues, we introduce a framework, which incorporates low-frequency perturbations to counteract face-swapping manipulation. This framework effectively attacks face-swapping models to protect images. Unlike perturbations in the spatial domain that tend to generate noise in detailed parts of the image—leading to the destruction of facial features and loss of details in other regions—our method preserves facial integrity while ensuring high image quality. This localized damage may be captured by the human eye, thus affecting the stealthiness of the attack. On the contrary, low-frequency perturbations in the frequency domain can adjust the facial features in general while preserving the details in the high-frequency part, making the image more visually natural. Therefore, we create a comprehensive framework for low-frequency perturbation against facial exchange models, consisting of an encoder-perturbation generator-decoder architecture. Specifically, spatial high-level characteristics are initially obtained from the input image using the encoder, with the DWT applied at the encoder's output to separate high-frequency and low-frequency components. Subsequently, we employ the perturbation generator to produce an imperceptible disturbance, and ultimately, the decoder reconstructs the image incorporating the perturbation at this stage, resulting in the modified image.

The key contributions of this research can be summarized as follows:
\begin{itemize}
\item This research recommends adding low-frequency perturbations to counteract face-swapping manipulation. Our method addresses the generation process of depth-faking techniques rather than influencing the classification results. We combine features from both the frequency and spatial domains to introduce artifacts through low-frequency perturbations while preserving high-frequency details, significantly reducing the generated content's naturalness and effectiveness while enhancing the overall quality of the altered images.
\item We design an integrated encoder-perturbation generator-decoder architecture to automate and seamlessly integrate perturbation by co-optimizing advanced feature extraction, perturbation generation, and reconstruction processes.
\item Extensive experiments show that the framework can effectively and successfully attack the facial exchange model and destroy the generated image to protect the image content.
\end{itemize}
\section{Related Work}
\subsection{Deepfake Generation}
With ongoing research in generative adversarial networks (GANs) \cite{b20}, deep-face forgery techniques have advanced, including face swapping, facial reconstruction, attribute editing, and face synthesis. Face swapping has become a key area of focus due to its applications in entertainment, movie production, and social media. This technique \cite{b6,b7,b8,b9,b11} transfers the source face's identity to the target face while preserving its attributes.
SimSwap \cite{b6} utilizes the Identity Injection Module (IIM) for identity transfer with feature matching, while InfoSwap \cite{b7} extracts identity information using a separation and exchange network. E4S \cite{b8} presents a high-fidelity approach by incorporating fine-grained face editing, and UniFace \cite{b9} introduces UCE loss to improve facial recognition model training. As these techniques evolve, the realism of generated images continues to increase.
\subsection{Deepfake Detection}
Facial swapping presents a significant threat to personal privacy and political security, necessitating effective countermeasures. L. Li et al. \cite{b21} proposed facial X-ray, a novel image representation for detecting facial forgery. Nirkin Y et al. \cite{b22} developed a method to detect face swapping and identity manipulation by analyzing the differences between a face and its surroundings using face recognition and contextual networks. B. Huang et al. \cite{b23} embedded facial images into explicit and implicit identity feature spaces, using the identity gap to assess authenticity. Chuangchuang Tan et al. \cite{b24} introduced Neighboring Pixel Relationships (NPR) to identify forgery artifacts from up-sampling operations. However, these methods are reactive and cannot prevent the spread of tampered images or the harm caused by false information.

To address after-the-fact forensics, recent research has focused on proactive defense by adding invisible information to images before uploading them to social media, aiming to prevent the generation and spread of false content. CMUA \cite{b14} generates image-specific cross-model perturbations targeting various Deepfake models. Smart Watermark \cite{b28} integrates imperceptible watermarks into the image's background, making Deepfake images appear indistinct, especially around the face. Huang Q et al. \cite{b16} proposed an active defense by introducing noise to facial data, reducing the manipulation model's effectiveness. Anti-forgery \cite{b17} enhances the resilience of perceptual-aware perturbations against distortions.

Meanwhile, Wang et al. \cite{b29} proposed DeepFake Disrupter, an efficient pipeline using generative networks to train perturbation generators that protect images or videos from various Deepfake manipulation techniques. These methods focus on attribute manipulation but show limited effectiveness against face swapping. TAFIM \cite{b18} is a data-driven strategy that safeguards face images by generating predefined manipulation targets when image manipulation is attempted. DF-RA \cite{b19} protects face images from Deepfake attacks in online social network compression environments, though its robustness against popular face-swapping models requires further validation. To address these issues, this paper introduces a framework that uses low-frequency perturbations to counter face-swapping manipulations, aiming to disrupt the attacking model and generate distorted, unrealistic images for enhanced protection.

\section{Method}
This part of the paper describes the specific structure and design elements of the proposed approach. Before delving into that, we define the problem.
\subsection{Problem Formulation}
\noindent\textbf{Facial Exchange.} We focus on attacking the face-swapping generative model, which derives the identity vector from the original image and incorporates the identity of the original image into the target image, all while retaining the characteristics of the target image. Thus, the objective of face-swapping can be articulated as:
\begin{equation}
y = G(x, x_t),
\label{eq:1}
\end{equation}
where \( G \) is the trained manipulation model, and \( y \) is the output image.
\begin{figure}[htbp]
\centerline{\includegraphics[width=0.5\textwidth]{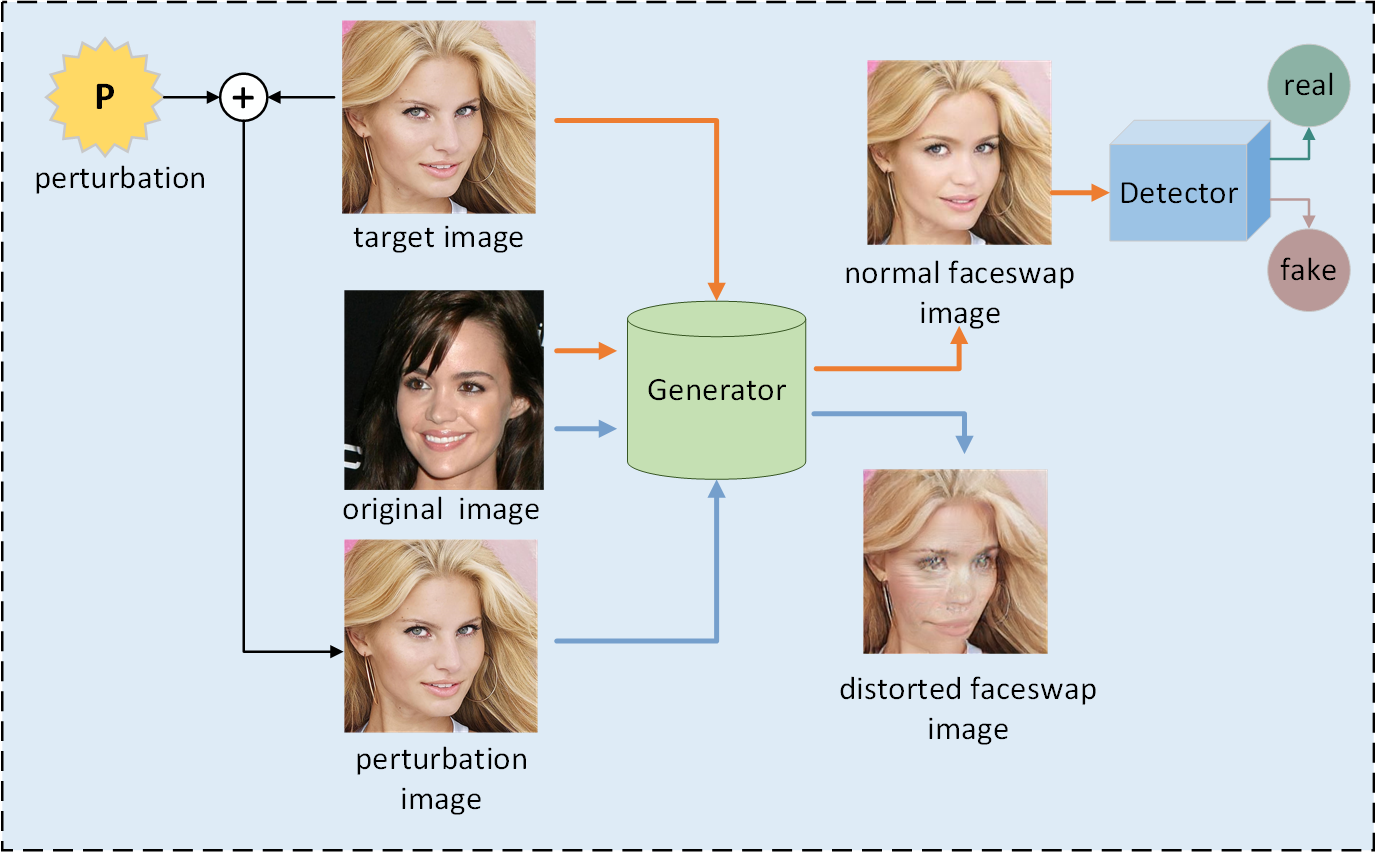}}
\caption{The orange line is the passive detection method, and the blue line is the active defense method.}
\label{fig:1}
\end{figure}

\noindent\textbf{Adversarial Attacks.} Active defense differs from passive detection in that passive detection detects the truth or falsity of false information after deep forgery. In contrast, active defense adds invisible information to the image before deep forgery to achieve protection of the image. We use a real example to illustrate passive detection and active defense, as shown in Fig.~\ref{fig:1}.

Since the face swap model requires two images as input, the perturbed image is utilized as the target image to generate the face swap image. Precisely similar to adversarial examples, we generate a perturbed image  $\tilde{x}_t$ by adding an imperceptible perturbation $\theta$ to the target image $x_t$:
\begin{equation}
\tilde{x}_t = x_t + \theta.
\label{eq:2}
\end{equation}

Specifically, \( y = G(x, x_t) \) and \( \tilde{y} = G(x, \tilde{x}_t) \) are further obtained by passing the original image \( x \) and the target image $x_t$ or the image $\tilde{x}_t$ containing the perturbation to the generative model \( G \). where \( y \) is the standard tampered image generated by the image \( x \) and the image through the generative model \( G \). \(\tilde{y}\) is the image with significant artifacts generated by the image \( x \) and the image $\tilde{x}_t$ containing the perturbation as the target image through the generative model \( G \).

To achieve the above effect, we maximize the distortion between \( y \) and \(\tilde{y}\) according to the expression below:
\begin{equation}
\max_{\theta} L(\tilde{y}, y), \quad \text{subject to } \|\theta\|_{\infty} \leq \epsilon,
\label{eq:3}
\end{equation}
where \(\epsilon\) represents the maximum perturbation magnitude, \( L \) denotes the distance function, and as the distance between \( y \) and \(\tilde{y}\) increases, the artifacts become more pronounced, enhancing the effectiveness of the attack.
\subsection{Method Overview}

\begin{figure*}[htbp] 
    \centering
    \centerline{\includegraphics[width=\textwidth]{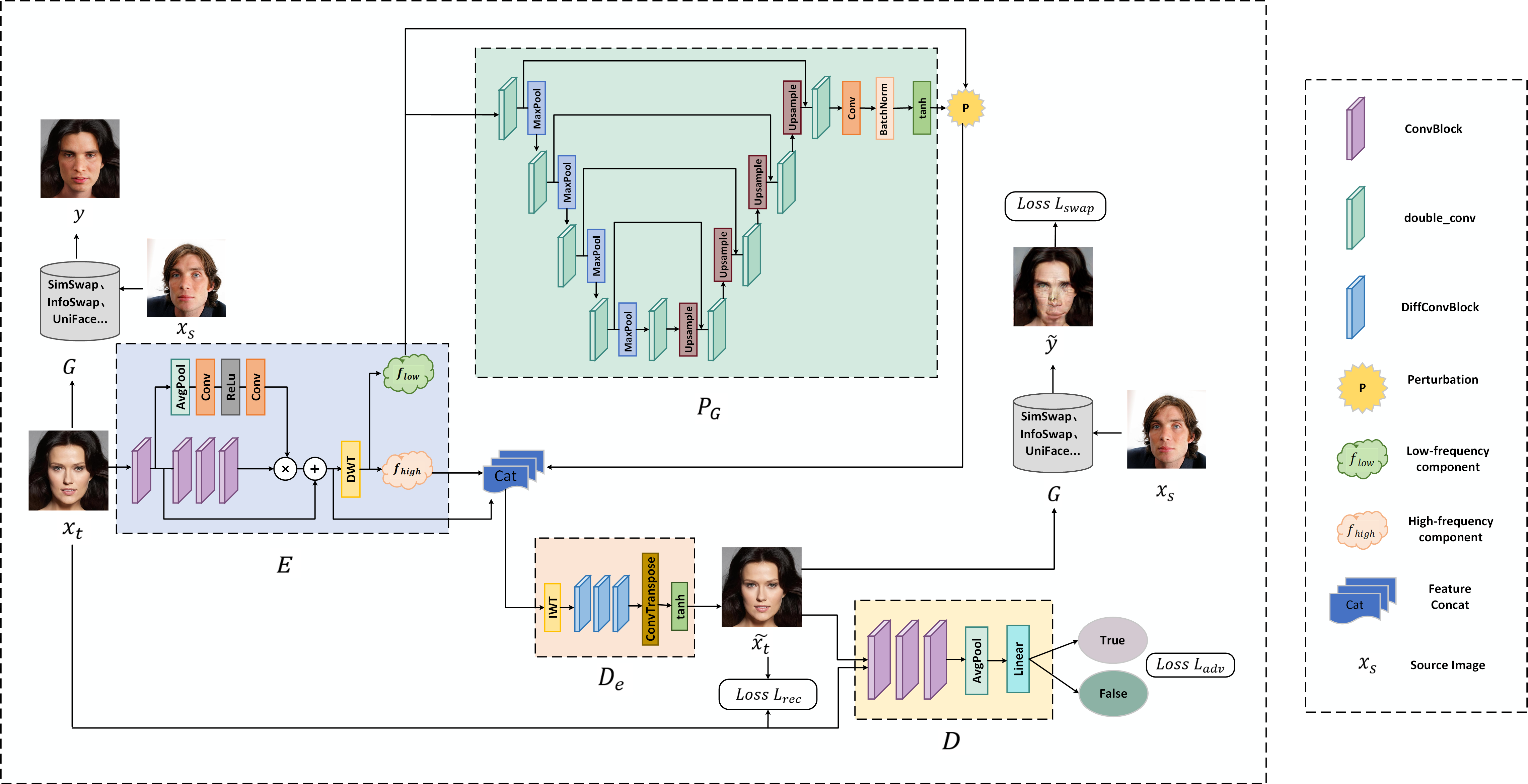}}
    \caption{presents the complete framework of our approach, highlighting the structure and process involved in applying low-frequency perceptual perturbations to combat facial manipulation attacks.} 
    \label{fig:2} 
\end{figure*}
Fig.~\ref{fig:2} shows the design of the model. Initially, the encoder \textit{E} extracts the abstract semantic features \( F_{\text{spatial}} \) from the target image $x_t$ and employs DWT to convert the image into the frequency domain, allowing the differentiation between the high-frequency component \( f_{\text{high}} \) and the low-frequency component \( f_{\text{low}} \). The perturbation generator $P_G$ takes the low-frequency component \( f_{\text{low}} \) as an input to generate the corresponding perturbation \textit{P}. The perturbation and the added perturbed low-frequency component \( f_{\text{low+p}} \), along with the high-frequency component \( f_{\text{high}} \) and spatial features \( F_{\text{spatial}} \), are sent to the decoder $D_e$ to reconstruct the perturbed image $x_t$. The scrambled image $\tilde{x}_t$ is not only visually similar to the original target image $x_t$ , but also generates a “false-at-a-glance” tampered image when the adversary maliciously manipulates it, thus blocking the dissemination of false content.
\subsection{Spatial-frequency Domain} 
We propose an encoder architecture that combines Discrete Wavelet Transform (DWT) \cite{b33} and spatial feature extraction to extract both the spatial and frequency domain information of an image simultaneously. The encoder's process involves spatial feature extraction, SE-ResNet enhanced features, and frequency domain decomposition.

\noindent\textbf{Spatial Feature Extraction.} First, given an input image $x_t$, we perform feature extraction using an initial convolution block. The convolution operation is defined as:
\begin{equation}
x_{\text{conv}} = \text{ConvBlock}\left(x_t\right),
\label{eq:4}
\end{equation}
where $\text{ConvBlock}$ denotes a standard convolution block. This process extracts the detailed features from the image's spatial dimension.

\noindent\textbf{SE-ResNet Enhanced Features.} We introduce a residual network module based on the attention mechanism after the initial convolutional block to further enhance the feature representation capability. This module utilizes the Squeeze-and-Excitation (SE) mechanism to compute channel-wise attention and preserves information flow using residual connections. Given the input feature \( x_{\text{conv}} \), the output through the residual module is represented as:
\begin{equation}
x_{\text{res}} = \text{SE-ResNet}\left(x_{\text{conv}}\right),
\label{eq:5}
\end{equation}
where SE-ResNet \cite{b30} denotes the residual network, which not only preserves spatial information but also models the relationship between channels through an adaptive attention mechanism. We further fuse the residual output with frequency domain features to preserve the critical structural information in the input image.

\noindent\textbf{Frequency Domain Decompositions.} After obtaining the spatial features, we utilize the discrete wavelet transform (DWT) to decompose the spatial features into the frequency domain. The wavelet transform aims to break down the feature map \( x_{\text{res}} \) into its low-frequency part \( x_{\text{LL}} \) and high-frequency parts \( x_{\text{HL}} \), \( x_{\text{LH}} \), \( x_{\text{HH}} \). Precisely, given the feature map \( x_{\text{res}} \), with the DWT, we can calculate:
\begin{equation}
f_{\text{low}}, f_{\text{high}} = \text{DWT}\left(x_{\text{res}}\right),
\label{eq:6}
\end{equation}
where \( f_{\text{high}} \) results from splicing the three high-frequency components \(x_{\text{HL}} \), \( x_{\text{LH}} \), and \( x_{\text{HH}} \), and \( f_{\text{low}} \) denotes the low-frequency component \( x_{\text{LL}} \). The standard wavelet transform operation is used here to decompose the frequency domain by downsampling and weighted summation.
\subsection{Perceived Perturbation} 
The perturbation generator proposed in this paper aims to produce the corresponding antagonistic perturbations based on the low-frequency details derived from the encoder:
\begin{equation}
P = P_{\text{G}}\left(f_{\text{low}}\right).
\label{eq:7}
\end{equation}

Specifically, the design of the generator is inspired by the symmetric jump-joining technique from U-Net \cite{b32}, allowing for the comprehensive utilization of feature information across various scales. Specifically, the generator consists of two stages: Downsampling and Upsampling. In the downsampling stage, the low-frequency components of the input are subjected to a series of convolution operations for feature extraction, gradually capturing global context information. 

Subsequently, the resolution of the feature map is reduced through an average pooling layer to capture more global features \( F_{\text{down}} \):
\begin{equation}
F_{\text{down}} = f_{\text{avgpool}}\big(f_{\text{conv}}(f_{\text{low}})\big),
\label{eq:8}
\end{equation}
where \( f_{\text{conv}} \) represents the convolution operation and \( f_{\text{avgpool}} \) represents the average pooling operation.

In the upsampling stage, the generator gradually restores the low-resolution feature map to a higher resolution through upsampling operations. At the same time, through jump connections, the features retained in the downsampling stage are spliced with the current upsampling features, so that global information and local details can be integrated. The purpose of this design is to retain detailed information while generating globally consistent perturbations. Ultimately, in this way, the generated adversarial perturbations can fully interfere with the image in the low-frequency domain without affecting the high-frequency details, thereby achieving a globally consistent perturbation effect.
\subsection{Reconstructed Image} 
The decoder design includes several key modules to restore high-quality perturbed images $\tilde{x}_t$:
\begin{equation}
\tilde{x}_t = D_e(f_{\text{low}}, f_{\text{high}}, P, F_{\text{spatial}}).
\label{eq:9}
\end{equation}

Specifically, the decoder concatenates the received low-frequency components \( f_{\text{low}} \), high-frequency components \( f_{\text{high}} \), and disturbances \textit{P} to obtain features \( F_{\text{cat}} \):
\begin{equation}
F_{\text{cat}} = f_{\text{low}} \oplus f_{\text{high}} \oplus P,
\label{eq:10}
\end{equation}
where \( \oplus \) represents concatenation. The convolution operation maps to 256 channel features. Two layers of convolution operations process the input, each followed by BatchNorm and ReLU activation functions to extract the image details. Then, the inverse wavelet transform (IWT) is applied to the mapped feature map to effectively combine low-frequency perturbations and high-frequency details, thereby enhancing the global information and details of the image.

On this basis, the decoder further integrates information from spatial features \( F_{\text{spatial}} \) and implements upsampling operations through layer-by-layer DiffConvBlock. DiffConvBlock replaces the traditional transposed convolution operation (ConvTranspose2d), effectively avoiding artifact problems and maintaining the naturalness of the image by gradually upsampling the resolution of the feature map. Finally, the decoder outputs the final generated image through the transposed convolution layer and the tanh activation function. This design ensures the generated image has high visual quality while retaining details and consistency.
\subsection{Objective Functions}
In the training phase, three loss functions are used: reconstruction loss, adversarial loss, and face-swapping loss.

\noindent\textbf{Reconstruction Loss.} The perturbation is added to the image $x_t$, and the objective is to maximize the resemblance of the perturbed image to the original one by applying a pixel-level $L_2$ constraint on the encoder. The reconstruction loss is defined as:
\begin{equation}
L_{\text{rec}} = \| \tilde{x}_t - x_t \|_2,
\label{eq:11}
\end{equation}
where $x_t$ denotes the unperturbed image, while $\tilde{x}_t$ indicates the version with the added perturbation.

\noindent\textbf{Adversarial Loss.} A discriminator \( D \) with parameter \( \theta_d \) is utilized to enhance the visual fidelity of the image, as described below:
\begin{equation}
L_{\text{adv}} = - \mathbb{E} \big( \log \big( D(\theta_d, x_t) \big) + \mathbb{E} \big( \log \big( 1 - D(\theta_d, \tilde{x}_t) \big) \big) \big).
\label{eq:12}
\end{equation}

The discriminator \( D \) is a binary classifier with three convolutional blocks that learns to recognize each pair $x_t$ and $\tilde{x}_t$ inversely during model supervision.

\noindent\textbf{Face-swap Loss.} The $L_1$ norm is used to bind the distortion between \( y \) and \(\tilde{y}\), as follows:
\begin{equation}
L_{\text{swap}} = - \| \tilde{y} - y \|_1,
\label{eq:13}
\end{equation}
where \( y \) is the output image obtained by applying the model \( G \) to the original image \( x \) and the target image $x_t$, and \(\tilde{y}\) is the result obtained by applying the model \( G \) to the original image \( x \) and the perturbed image $\tilde{x}_t$. The goal is to make the gap between these two as large as possible.

\noindent\textbf{Total Loss.} The overall loss is formulated as a weighted sum of three loss components:
\begin{equation}
L_{\text{total}} = \lambda_1 L_{\text{rec}} + \lambda_2 L_{\text{adv}} + \lambda_3 L_{\text{swap}},
\label{eq:14}
\end{equation}
where \( \lambda_1 \), \( \lambda_2 \), \( \lambda_3 \) are the weights that determine the influence of each loss component.

\section{Experiments}
\subsection{Experimental Details}

\noindent\textbf{Dataset.} We experimented with the high-resolution CelebA-HQ dataset \cite{b36}, which includes 30,000 image samples and features 6,217 unique identities. The standard division for training, validation, and testing \cite{b37} was adhered to. The trained model was tested on the LFW dataset \cite{b38}, comprising 5,749 distinct identities, to further validate its generalization ability. The framework was trained end-to-end on 2 Tesla V100 GPUs.

\noindent\textbf{Parameters.} The learning rates of both our framework and discriminator were configured to 0.0001. Empirically, we adopted $\lambda_1 = 5.0$, $\lambda_2 = 0.05$, and $\lambda_3 = 1.0$ in the relevant loss functions.

\noindent\textbf{Evaluation Metrics.} In this study, the visual quality was assessed using the mean peak signal-to-noise ratio (PSNR) \cite{b39} and the structural similarity index (SSIM) \cite{b40}. Since face-swapping damages are more dispersed than those in attribute editing, the average norm distance between the forged and original faces was calculated to measure degradation. The defense was considered successful if the distance exceeded 0.05. Based on this, the attack success rate (ASR) was defined as the ratio of forged images within the perturbed set that were effectively disrupted. The best performance is shown in bold in all tables.
\begin{figure}[htbp]
\centerline{\includegraphics[width=0.5\textwidth]{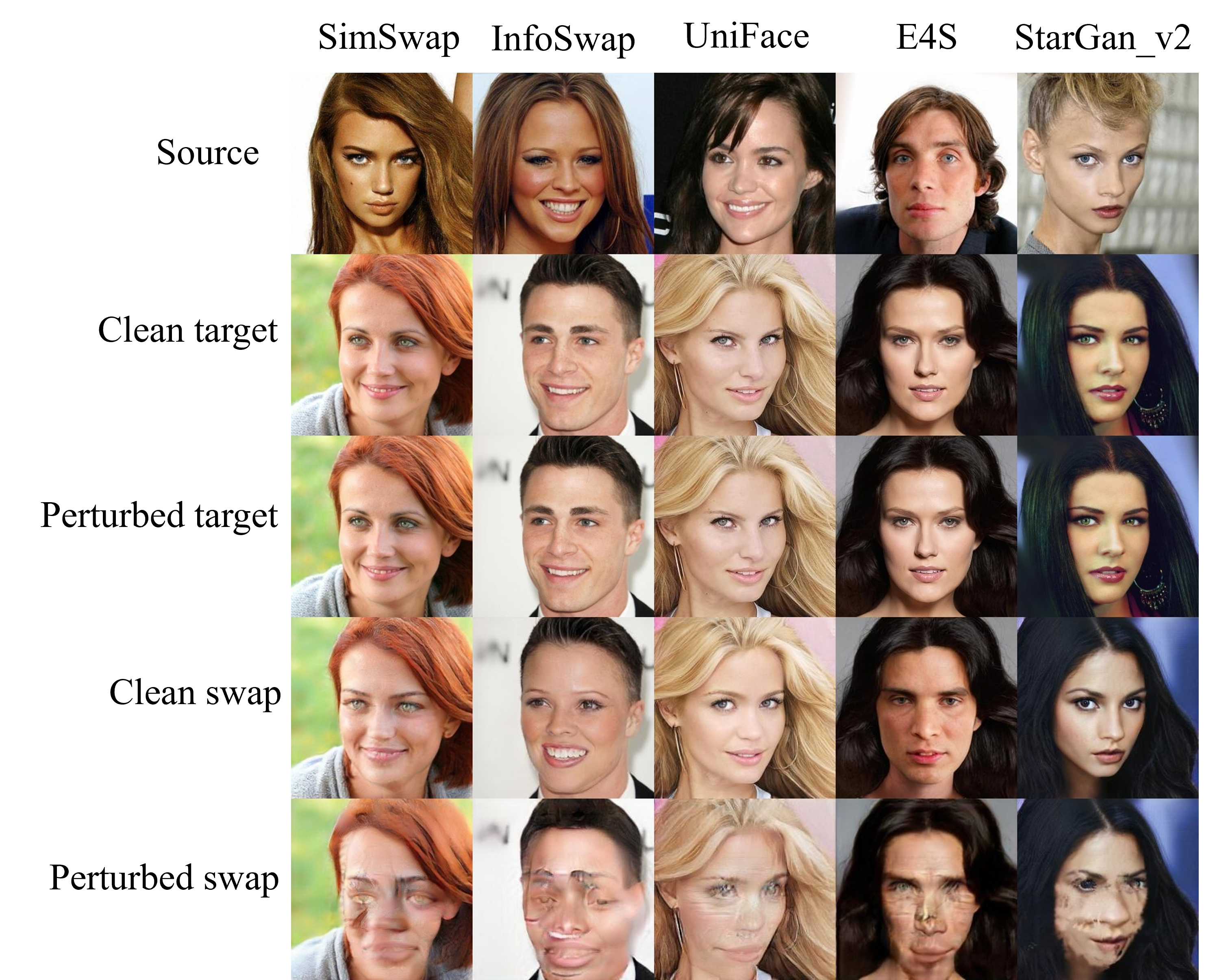}}
\caption{The first row is the source image input to the tampering model, the second row is the clean target image (without perturbation), the third row is the target image after the perturbation, the fourth row is the result image generated by the source image and the clean target image, and the fifth row is the result image generated by the source image and the target image with the perturbation.}
\label{fig:3}
\end{figure}
\begin{table}[ht]
\centering
\caption{Quantitative visual quality evaluation of perturbed images on the CelebA-HQ dataset. Information includes model name, PSNR, and SSIM.}
\begin{tabular}{lccccl}
\toprule
\textbf{Model} & \textbf{PSNR$\uparrow$} & \textbf{SSIM$\uparrow$} \\
\midrule
CMUA \cite{b14}    & 38.46 & 0.843 \\
Anti-Forgery \cite{b17}     & 35.75 & 0.937 \\
DF-RAP \cite{b19}       & 36.95 & 0.885 \\
IDFM \cite{b16}       & 33.94 & 0.734 \\
Disrupting Deepfakes \cite{b15}    & 35.53 & 0.759 \\
\textbf{Ours}  & \textbf{38.93} & \textbf{0.940} \\
\bottomrule
\end{tabular}
\label{tab:1}
\end{table}
\begin{table*}[ht]
\centering
\caption{Qualitative evaluation of the mean $L_1$ norm distance between \( y \) and \(\tilde{y}\) on the CelebA-HQ dataset, which includes the names of the comparison models and those of the generated manipulation models.}
\begin{tabular}{lcccccc}
\toprule
\textbf{Model} & \textbf{CMUA \cite{b14}} & \textbf{Anti-Forgery \cite{b17}} & \textbf{DF-RAP \cite{b19}} & \textbf{IDFM \cite{b16}} & \textbf{Disrupting Deepfakes \cite{b15}} & \textbf{Ours} \\
\midrule
SimSwap \cite{b6}    & 0.0079 & 0.0165 & \textbf{0.0495} & 0.0187 & 0.0134 & 0.0412 \\
InfoSwap \cite{b7}   & 0.0098 & 0.0432 & 0.0121 & 0.0446 & 0.0380 & \textbf{0.0463} \\
UniFace \cite{b9}    & 0.0136 & 0.0168 & 0.0078 & 0.0138 & 0.0113 & \textbf{0.0341} \\
E4S \cite{b8}        & 0.0046 & 0.0077 & 0.0146 & 0.0184 & 0.0067 & \textbf{0.0387} \\
StarGan\_v2 \cite{b41}  & 0.0473 & 0.0486 & 0.0436 & 0.0535 & 0.0514 & \textbf{0.0557} \\
\bottomrule
\end{tabular}
\label{tab:2}
\end{table*}
\begin{table*}[ht]
\centering
\caption{Qualitative assessment of the defense success rate on the CelebA-HQ dataset, including the names of the comparison models and the generated manipulated models.}
\begin{tabular}{lcccccc}
\toprule
\textbf{Model} & \textbf{CMUA \cite{b14}} & \textbf{Anti-Forgery \cite{b17}} & \textbf{DF-RAP \cite{b19}} & \textbf{IDFM \cite{b16}} & \textbf{Disrupting Deepfakes \cite{b15}} & \textbf{Ours} \\
\midrule
SimSwap \cite{b6}    & 0.09\% & 0.24\% & \textbf{98.92\%} & 0.16\% & 0.08\% & 80.65\% \\
InfoSwap \cite{b7}   & 1.59\% & 22.95\% & 0.60\% & 62.63\% & 16\% & \textbf{82.62\%} \\
UniFace \cite{b9}    & 3.68\% & 0.17\% & 0.18\% & 0.09\% & 0.13\% & \textbf{40.35\%} \\
E4S \cite{b8}        & 0.05\% & 0.10\% & 1.52\% & 3.90\% & 0.09\% & \textbf{49.68\%} \\
StarGan\_v2 \cite{b41}  & 54.89\% & 35.89\% & 46.28\% & 48\% & 40.15\% & \textbf{87.79\%} \\
\bottomrule
\end{tabular}
\label{tab:3}
\end{table*}
\begin{table}[ht]
\centering
\caption{Quantitative visual quality evaluation of perturbed images on the LFW dataset. Information includes model name, PSNR, and SSIM.}
\begin{tabular}{lccccl}
\toprule
\textbf{Model} & \textbf{PSNR$\uparrow$} & \textbf{SSIM$\uparrow$} \\
\midrule
CMUA \cite{b14}    & 38.62 & 0.862 \\
Anti-Forgery \cite{b17}     & 36.98 & 0.943 \\
DF-RAP \cite{b19}       & 38.06 & 0.905 \\
IDFM \cite{b16}       & 34.59 & 0.729 \\
Disrupting Deepfakes \cite{b15}    & 34.57 & 0.746 \\
\textbf{Ours}  & \textbf{39.06} & \textbf{0.947} \\
\bottomrule
\end{tabular}
\label{tab:4}
\end{table}
\begin{table*}[ht]
\centering
\caption{Qualitative evaluation of the average norm distance between \( y \) and \(\tilde{y}\) on the LFW dataset, including the names of the comparison models and the generated manipulation models.}
\begin{tabular}{lcccccc}
\toprule
\textbf{Model} & \textbf{CMUA \cite{b14}} & \textbf{Anti-Forgery \cite{b17}} & \textbf{DF-RAP \cite{b19}} & \textbf{IDFM \cite{b16}} & \textbf{Disrupting Deepfakes \cite{b15}} & \textbf{Ours} \\
\midrule
SimSwap \cite{b6}    & 0.0076 & 0.0166 & \textbf{0.0474} & 0.0185 & 0.0145 & 0.0407 \\
InfoSwap \cite{b7}   & 0.0085 & 0.0426 & 0.0116 & 0.0442 & 0.0380 & \textbf{0.0469} \\
UniFace \cite{b9}    & 0.0119 & 0.0157 & 0.0079 & 0.0140 & 0.0098 & \textbf{0.0343} \\
E4S \cite{b8}        & 0.0059 & 0.0074 & 0.0150 & 0.0205 & 0.0096 & \textbf{0.0381} \\
StarGan\_v2 \cite{b41}  & 0.0480 & 0.0492 & 0.0428 & 0.0476 & 0.0537 & \textbf{0.0549} \\
\bottomrule
\end{tabular}
\label{tab:5}
\end{table*}
\begin{table*}[ht]
\centering
\caption{Qualitative evaluation of defense success rate on the LFW dataset, including the names of the comparison models and the generated manipulation models.}

\begin{tabular}{lcccccc}
\toprule
\textbf{Model} & \textbf{CMUA \cite{b14}} & \textbf{Anti-Forgery \cite{b17}} & \textbf{DF-RAP \cite{b19}} & \textbf{IDFM \cite{b16}} & \textbf{Disrupting Deepfakes \cite{b15}} & \textbf{Ours} \\
\midrule
SimSwap \cite{b6}    & 0.10\% & 0.23\% & \textbf{97.64\%} & 0.14\% & 0.09\% & 80.43\% \\
InfoSwap \cite{b7}   & 1.61\% & 22.57\% & 0.54\% & 62.93\% & 15.74\% & \textbf{83.77\%} \\
UniFace \cite{b9}    & 2.14\% & 0.15\% & 0.17\% & 0.13\% & 0.11\% & \textbf{40.28\%} \\
E4S \cite{b8}        & 0.06\% & 0.12\% & 1.49\% & 4.16\% & 0.18\% & \textbf{50.20\%} \\
StarGan\_v2 \cite{b41}  & 56.14\% & 36.34\% & 47.36\% & 47.86\% & 41.06\% & \textbf{87.36\%} \\
\bottomrule
\end{tabular}
\label{tab:6}
\end{table*}

\begin{table}[ht]
\centering
\caption{This table shows the visual quality (PSNR and SSIM) of the perturbed images and the defense success rate (ASR) at different perturbation amplitudes.}
\begin{tabular}{lcccccc}
\toprule
\textbf{Metrics} & \textbf{0.01} & \textbf{0.02} & \textbf{0.03} & \textbf{0.04} & \textbf{0.05} & \textbf{0.06} \\
\midrule
PSNR   & 42.68 & 42.15 & 40.56 & 39.96 & 38.93 & 36.77 \\
SSIM   & 0.994 & 0.987 & 0.964 & 0.952& 0.940 & 0.928 \\
ASR    & 0.09\% & 0.17\% & 23.59\% & 50.64\% & 80.65\% & 83.58\% \\
\bottomrule
\end{tabular}
\label{tab:7}
\end{table}

\subsection{Experiment on CelebA-HQ}
We assessed the impact of the proposed technique on CelebA-HQ by measuring visual quality, the mean $L_1$ norm difference between \( y \) and \(\tilde{y}\), and the defense success rate (ASR).The comparison methods included adversarial attack models (CMUA \cite{b14}, Anti-Forgery \cite{b17}, DF-RAP \cite{b19}, IDFM \cite{b16}, Disrupting Deepfakes \cite{b15}). We directly adopted the publicly available training model weights corresponding to the best performance. Due to the unavailability of the 2023 model source code and data, this study failed to compare the 2023 algorithms or results.

\noindent\textbf{Visual Quality.} Fig.~\ref{fig:3} displays the sampled images. Specifically, we visualized the original image \( x \), target image $x_t$, perturbed image $\tilde{x}_t$, face-swapped image \( y \) created using the original and target images, and face-swapped image \(\tilde{y}\) created using the original and perturbed images, listed from top to bottom. Additionally, the third and fourth rows show, from left to right, images processed by the SimSwap \cite{b6}, InfoSwap \cite{b7}, UniFace \cite{b9}, E4S \cite{b8}, and StarGan\_v2 \cite{b41} models. In order to assess the performance of our approach in face reconstruction, we also performed attack tests on the classic StarGan\_v2 \cite{b41} model.

It can be seen from Fig.~\ref{fig:3} that the perturbed image $\tilde{x}_t$ is visually comparable to the original image $x_t$, while the face-swapped image \(\tilde{y}\) exhibits obvious distortion and deformation. This demonstrates that our perturbations can successfully interfere with the attack generation model without compromising the visual integrity of the image.

In the quantitative analysis presented in Table~\ref{tab:1}, we computed the mean peak signal-to-noise ratio (PSNR) and the structural similarity index (SSIM) between the original and perturbed images. These metrics assess the degree of perturbation and structural similarity of our framework. Table~\ref{tab:1} shows that our approach excels in preserving visual quality, outperforming prior methods. The higher PSNR and SSIM values indicate that the perturbation's effect on the original image is nearly undetectable. CMUA \cite{b14} and DF-RAP \cite{b19} also exhibit good visual quality, with CMUA \cite{b14} achieving the second-best result in PSNR.

\noindent\textbf{\boldmath$L_1$ Norm Distance.} Table~\ref{tab:2} demonstrates that our approach achieves high $L_1$ values across various face-swapping and expression manipulation models, indicating exceptional performance. For example, on InfoSwap \cite{b7}, our method reaches 0.0463, significantly outperforming other anti-counterfeiting strategies. Additionally, it achieves optimal or near-optimal values on SimSwap \cite{b6} and StarGan\_v2 \cite{b41}, showing that our method is more effective in generating significant perturbations that disrupt the image generation process. In contrast, while DF-RAP \cite{b19} performs well on some algorithms (e.g., SimSwap \cite{b6}), our approach consistently achieves higher $L_1$ values across multiple evaluation criteria. These results highlight the effectiveness of our method in resisting face-swapping attacks and generating adversarial perturbations.

\noindent\textbf{Defense Success Rate.} Table~\ref{tab:3} shows that our approach achieves high defense success rates across most face-swapping algorithms. For instance, on SimSwap \cite{b6} and InfoSwap \cite{b7}, our method reaches success rates of 80.65\% and 82.62\%, respectively, outperforming other anti-counterfeiting methods. It also achieves 87.79\% on the StarGan\_v2 \cite{b41}. These results demonstrate the robustness of our approach in resisting attacks across different face-replacement scenarios. In contrast, while DF-RAP \cite{b19} performs well on SimSwap \cite{b6} (98.92\%), its performance is less stable on other models. Methods like CMUA \cite{b14}, Anti-Forgery \cite{b17}, and Disrupting Deepfakes \cite{b15} show generally low success rates, suggesting room for improvement in attack resistance. Overall, our method shows significant advantages in attack success rate, confirming its broad applicability and efficiency in anti-counterfeiting tasks.

\subsection{Cross-Dataset Experiment}
We further evaluate the generalization ability of our framework on an unseen dataset, Labeled Faces in the Wild (LFW) \cite{b38}. This evaluation tests whether our proposed method can maintain robust performance and good generalization when facing unknown datasets. As shown in Table~\ref{tab:4}, our model performs strongly compared to alternative methods, particularly in terms of image quality and structural fidelity. Our model more effectively preserves the visual details and structure of the original image.
\begin{figure}[htbp]
\centerline{\includegraphics[width=0.5\textwidth]{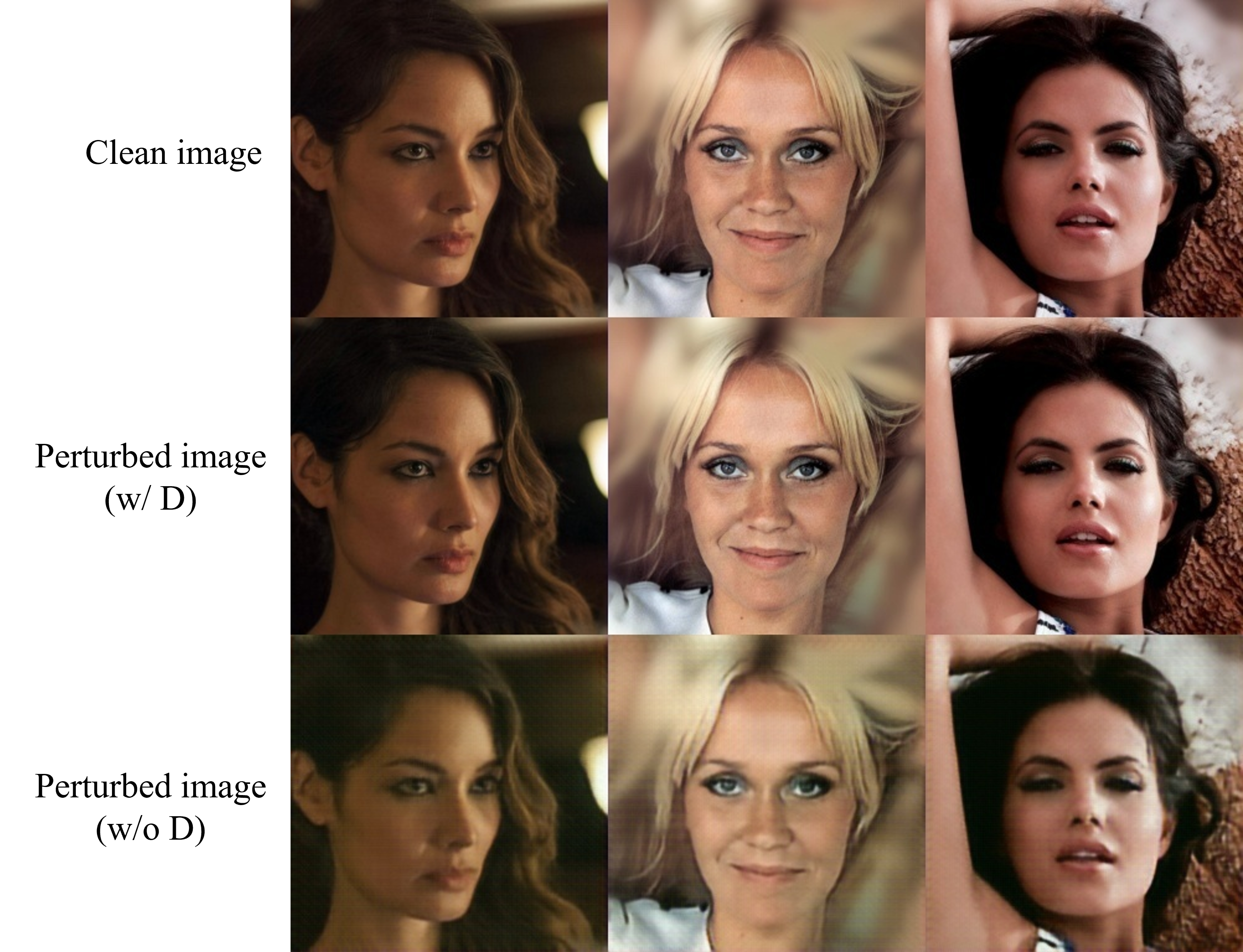}}
\caption{The first row is the clean image, the second row is the perturbed image with the discriminator, and the third row is the perturbed image without the discriminator.}
\label{fig:4}
\end{figure}

\noindent\textbf{\boldmath$L_1$ Norm Distance.} By analyzing the data in Table~\ref{tab:5}, we observe that our model performs better than other defense methods in handling various face-swapping operations and demonstrates a higher attack-blocking effect. While other methods exhibit specific performance in individual face-swapping algorithms, their overall performance is not as stable and excellent as our model's. This result highlights the generalization ability and robustness of our method, proving that it has practical application potential and broad promotion value in face-swapping manipulation defense.

\noindent\textbf{Defense Success Rate.} As shown in Table~\ref{tab:6}, our method achieves a high attack success rate in various face replacement algorithms, such as 80.43\% on SimSwap \cite{b6}, 83.77\% on InfoSwap \cite{b7}, and 87.36\% on StarGan\_v2 \cite{b41}, outperforming other defense strategies. In contrast, DF-RAP \cite{b19} achieved 97.64\% on SimSwap but only 0.54\% on InfoSwap. Other methods like CMUA \cite{b14}, Anti-Forgery \cite{b17}, and Disrupting Deepfakes \cite{b15} showed weaker results, with CMUA \cite{b14} reaching just 56.14\% on StarGan\_v2 \cite{b41} and 0.10\% on SimSwap \cite{b6}, indicating their limited effectiveness. Overall, our method demonstrates significant advantages in attack success rates and stability, proving its robustness in anti-counterfeiting applications.

\subsection{Ablation Studies}
\noindent\textbf{Discriminator Selection.} In Fig.~\ref{fig:4}, we show the visual quality of the perturbed images with and without the discriminator in the entire framework. The visual quality of the perturbed images obtained with the discriminator is much higher than that without the discriminator. Therefore, we retain the discriminator when training the entire framework.

\noindent\textbf{Disturbance Range.} 
From the three perspectives of PSNR, SSIM, and ASR in Table~\ref{tab:7}, we finally choose a perturbation amplitude of 0.05 to balance image quality and attack success rate. For example, SimSwap on CelebA-HQ shows a PSNR of 38.93 at this amplitude, which, while lower than smaller perturbations, still preserves high image quality. The SSIM is 0.940, indicating good structural preservation, compared to 0.928 at a higher amplitude of 0.06. In terms of ASR, the rate is 80.65\% at 0.05, higher than smaller amplitudes, and only marginally lower than 83.58\% at 0.06. Therefore, a perturbation amplitude of 0.05 achieves a high attack success rate while maintaining image quality and structure.
\section{Conclusion}
This study introduces a framework that uses low-frequency perturbations to counteract face-swapping manipulations in deepfake technology. By applying perturbations to the low-frequency regions of the frequency spectrum, we reduce the impact of attack models on face manipulation. The framework includes an encoder, a perturbation generator, and a decoder. The encoder extracts high- and low-frequency components and spatial features using discrete wavelet transform (DWT). The perturbation generator creates invisible perturbations based on low-frequency components and applies them accordingly. The decoder then reconstructs the image using inverse wavelet transform (IWT). This method preserves high-frequency details while ensuring the imperceptibility of perturbations, enhancing defense against face-swapping attacks. Experimental results show that the framework performs well across datasets (CelebA-HQ and LFW), effectively disrupting face-swapping models while maintaining image quality. Our approach outperforms existing defense strategies in attack success rate and defense generalization, demonstrating robustness across different scenarios. However, the performance gap from perfect detection points to future research directions, focusing on improving generalizability, especially against face reenactment manipulations.

\section*{Acknowledgment}
This study was supported by the Key R \& D Program of Shandong Province (Major Scientific and Technological Innovation Projects) under Grant Nos. 2024CXGC010109, 2023CXGC010113, and 2024CXGC010101; the "20 New Universities" Program of Jinan under Grant No. 202228068; and the Major Innovation Project of the Science-Education-Industry Integration Pilot Program of Qilu University of Technology (Shandong Academy of Sciences) under Grant No. 2024ZDZX11. The authors sincerely thank the editor and anonymous reviewers for their insightful comments, which greatly helped us to improve the quality of this paper.

\end{document}